\documentclass[sigconf]{acmart}

\AtBeginDocument{}
\setcopyright{none}
\renewcommand\footnotetextcopyrightpermission[1]{}
\acmConference[WWW '27]{The ACM Web Conference 2027}{2027}{Dublin, Ireland}
\acmBooktitle{Proceedings of The ACM Web Conference 2027 (WWW '27)}
\acmYear{2027}
\acmDOI{}
\acmISBN{}

\usepackage{amsmath,booktabs,multirow,array}
\usepackage{colortbl,tabularx}
\definecolor{BordInk}{HTML}{283644}
\definecolor{BordTeal}{HTML}{006B63}
\definecolor{BordHeader}{HTML}{EDF1F5}
\definecolor{BordGroup}{HTML}{F5F7F9}
\definecolor{BordTint}{HTML}{E8F3F0}
\definecolor{BordRule}{HTML}{718291}
\newcommand{\bordtablestyle}{\renewcommand{\arraystretch}{1.18}\setlength{\tabcolsep}{3pt}\setlength{\heavyrulewidth}{0.7pt}\setlength{\lightrulewidth}{0.35pt}\arrayrulecolor{black}}
\newcommand{\bordhead}[1]{\textcolor{BordInk}{\textbf{#1}}}
\newcommand{\bordname}[1]{\textcolor{BordTeal}{\textbf{#1}}}
\newcommand{\best}[1]{\textbf{#1}}
\newcommand{\second}[1]{\underline{#1}}
\usepackage{microtype}
\usepackage{algorithm}
\usepackage{algpseudocode}
\usepackage{etoolbox}
\usepackage{float}
\makeatletter
\patchcmd{\ACM@mk@linecount}{\color{red}}{\color{black}}{}{}
\patchcmd{\ACM@mk@linecount}{\color{red}}{\color{black}}{}{}
\makeatother

\newcommand{\method}{SA-OPSD}

\newcommand{\beam}{\mathcal{B}}
\newcommand{\frontier}{\mathcal{F}}
\newcommand{\lossrel}{\mathcal{L}_{\mathrm{rel}}}
\newcommand{\losscond}{\mathcal{L}_{\mathrm{cond}}}
\newcommand{\lossmass}{\mathcal{L}_{\mathrm{mass}}}
\newcommand{\lossfull}{\mathcal{L}_{\mathrm{full}}}

\title[SA-OPSD]{Search-Aware On-Policy Self-Distillation with Sibling Alignment for Generative Recommendation}

\author{\texorpdfstring{\href{https://openreview.net/profile?id=~Rongfeng_Guo1}{Rongfeng Guo}\textsuperscript{1}}{Rongfeng Guo}}
\affiliation{\institution{Kuaishou Technology}\city{Beijing}\country{China}}

\author{\texorpdfstring{\href{https://openreview.net/profile?id=~Zhuang_Zhuang1}{Zhuang Zhuang}\textsuperscript{1}}{Zhuang Zhuang}}
\affiliation{\institution{Kuaishou Technology}\city{Beijing}\country{China}}

\author{\texorpdfstring{\href{https://openreview.net/profile?id=~Ji_Dai2}{Ji Dai}\textsuperscript{1}}{Ji Dai}}
\affiliation{\institution{Kuaishou Technology}\city{Beijing}\country{China}}

\author{\texorpdfstring{\href{https://openreview.net/profile?id=~Peng_Zhao26}{Peng Zhao}\textsuperscript{1}}{Peng Zhao}}
\affiliation{\institution{Kuaishou Technology}\city{Beijing}\country{China}}

\author{\texorpdfstring{\href{https://openreview.net/profile?id=~Shijie_Li24}{Shijie Li}\textsuperscript{1}}{Shijie Li}}
\affiliation{\institution{Kuaishou Technology}\city{Beijing}\country{China}}

\author{\texorpdfstring{\href{https://openreview.net/profile?id=~Jie_Chen61}{Jie Chen}\textsuperscript{1}}{Jie Chen}}
\affiliation{\institution{Kuaishou Technology}\city{Beijing}\country{China}}

\author{\texorpdfstring{\href{https://openreview.net/profile?id=~Teng_Long2}{Teng Long}\textsuperscript{3}}{Teng Long}}
\affiliation{\institution{University of Amsterdam}\city{Amsterdam}\country{Netherlands}}

\author{\texorpdfstring{\href{https://openreview.net/profile?id=~Vincent_Tao_Hu1}{Vincent Tao Hu}\textsuperscript{2}}{Vincent Tao Hu}}
\affiliation{\institution{Huazhong University of Science and Technology}\city{Wuhan}\country{China}}

\author{\texorpdfstring{\href{https://openreview.net/profile?id=~Fei_Pan1}{Fei Pan}\textsuperscript{1}}{Fei Pan}}
\affiliation{\institution{Kuaishou Technology}\city{Beijing}\country{China}}

\AtEndDocument{\nobalance}

\begin{document}

\begin{abstract}
Generative recommendation retrieves items by autoregressively generating semantic IDs, typically via beam search.
On-policy self-distillation (OPSD) enhances training supervision by transferring privileged self-teacher knowledge at student-generated states.
However, beam-search retention depends on competition among prefixes: visited states differ in importance, and initially reliable decisions may approach the pruning boundary as training evolves.
Moreover, conventional distribution matching couples target-token confidence with preferences among competing continuations, potentially weakening relational supervision when the self-teacher strongly favors the target.
To address these challenges, we propose \textbf{Search-Aware On-Policy Self-Distillation (SA-OPSD)}, a framework that transfers privileged self-teacher knowledge to the states visited during the student's own beam search.
Specifically, we introduce \textbf{Evolving-Boundary Sampling}, which prioritizes search-critical states by considering both current pruning-boundary proximity and initial-reference retention-margin decline under a controlled distillation budget.
Furthermore, we develop \textbf{Confidence-Decoupled Branch Distillation}, which separates target-probability calibration from informative sibling relations to better preserve the teacher's preferences among competing branches.
Experiments on Industrial, Office, and Instruments demonstrate strong performance for SA-OPSD.
\end{abstract}

\maketitle
\begin{figure}[t]
\centering
\includegraphics[width=0.98\columnwidth]{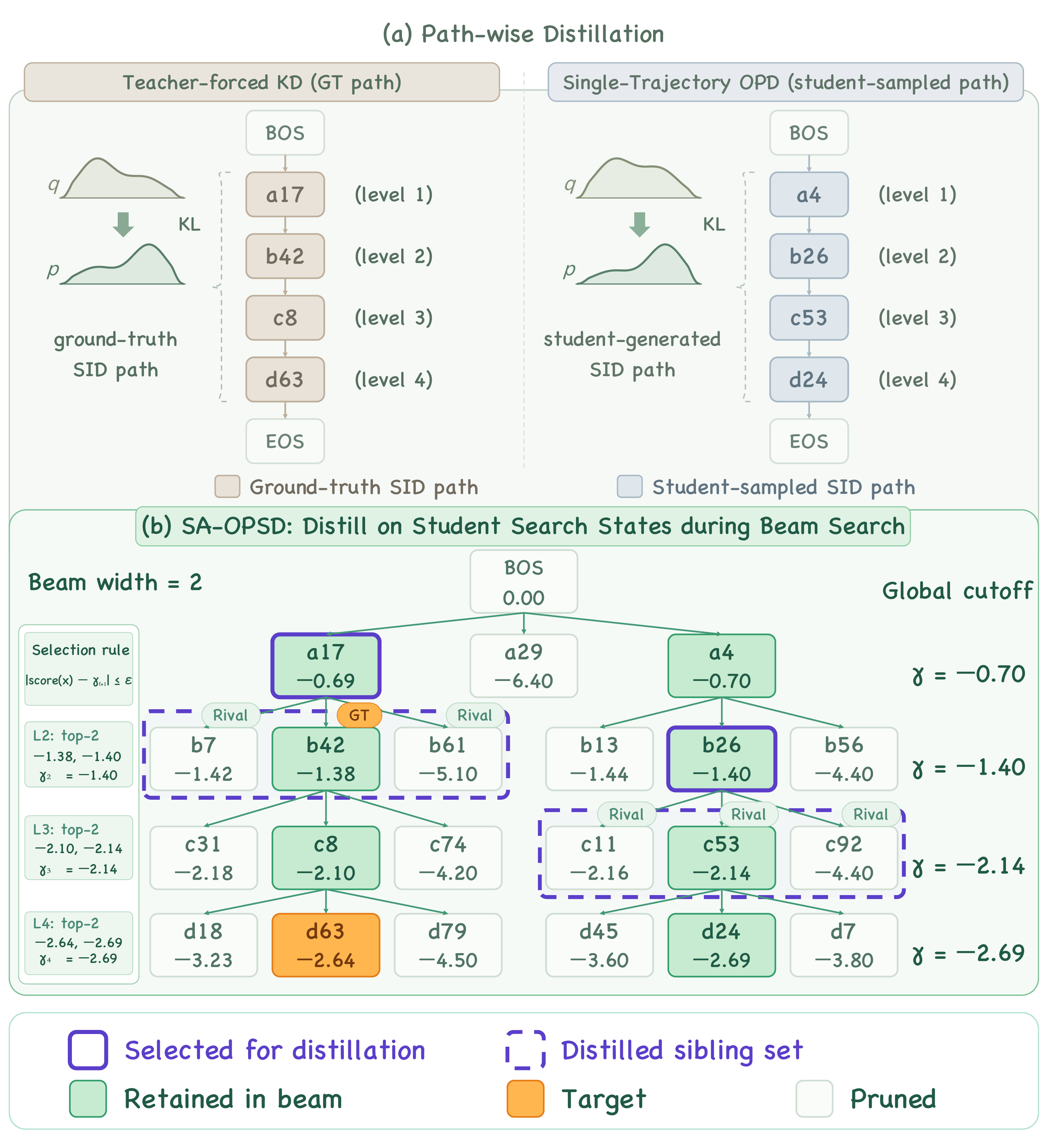}
\caption{Path-wise and student beam-search distillation. (a) Two illustrative path-wise settings supervise the ground-truth SID path or a single student-sampled path. (b) SA-OPSD selects GT-prefix and non-GT-prefix parents from competing beam-search states. Scores and highlighted parents are illustrative. EBS sampling is defined in Sec.~\ref{sec:rbs-budget}.}
\Description{Two-part comparison of path-wise distillation and student beam-search states. The left part displays target-path and single-trajectory supervision; the right part shows beam width two, cumulative prefix scores, retained and pruned candidates, and selected parent states.}
\label{fig:motivation}
\end{figure}
\section{Introduction}

Generative recommendation (GR) formulates next-item prediction as the autoregressive generation of semantic ID (SID) tokens conditioned on a user's interaction history~\cite{rajput2023tiger,zheng2023lcrec,wang2024letter}. Knowledge distillation (KD) transfers preference knowledge to student recommenders during training~\cite{hinton2015distilling,tang2018ranking,cui2024dllm2rec,guo2026sidmlp,zhang2026smartgr}. Here, we study on-policy self-distillation for students deployed with beam search, keeping their architecture and decoding procedure unchanged. Beam search expands candidate SID prefixes and retains the highest-scoring ones at each decoding step~\cite{yu2026apao}. In Web-based product discovery, generative recommenders must efficiently retrieve and rank relevant products from large catalogs. Since beam-search pruning determines which candidates remain available for ranking, distillation should reflect the student's actual decoding decisions.

Knowledge distillation has been extensively investigated in recommendation, from ranking-based knowledge transfer~\cite{tang2018ranking,kang2023hetcomp} to LLM-based recommendation distillation~\cite{cui2024dllm2rec,chen2025c2kd} and generative recommendation models~\cite{zhang2026smartgr}. Beyond conventional teacher-forcing supervision, on-policy distillation (OPD) transfers teacher knowledge at states induced by the student~\cite{agarwal2023gkd}. Recent recommendation research has also explored on-policy distillation for latent reasoning~\cite{zhuang2026evoreason}. Meanwhile, self-distillation provides another direction for exploiting richer supervision, with privileged information from future interactions or reasoning contexts guiding recommendation models~\cite{tang2026psd,jiang2026selfdr}. On-policy self-distillation (OPSD) further combines student-generated trajectories with privileged self-teacher supervision~\cite{zhao2026opsd}. Related recommendation research also explores student-aware reasoning distillation~\cite{shahgir2026scored}. These advances offer a promising foundation for transferring richer preference knowledge to generative recommenders.

On-policy distillation grounds supervision in states generated by the student, making the rollout procedure central to knowledge transfer. For beam-decoded recommendation, the student's own beam-search rollouts expose competing prefixes whose continuations contend for shared beam slots. Candidates are retained according to their cumulative path scores, and a target prefix may be pruned even when its local continuation probability is relatively high. Moreover, changes to competing continuations can alter a target's rank even when its own score is fixed. Recent studies have explicitly incorporated such search behavior into recommendation learning. APAO~\cite{yu2026apao} optimizes target-prefix survival, BEAR~\cite{yang2026bear} introduces beam-search-aware regularization, and SmartGR~\cite{zhang2026smartgr} transfers teacher-beam ranking preferences. These studies highlight the importance of search-level optimization. Consistent with this motivation, the rollout comparison in Table~\ref{tab:rollout-strategies} shows higher NDCG@10 for dense beam-search supervision than for single- and multi-rollout distillation, motivating supervision over competing student-visited states.

Although OPSD enables supervision over student-generated states, directly extending it to beam-search prefixes introduces two coupled challenges. First, \emph{where should a limited distillation budget be spent?} Student beam search produces numerous parent states with different influences on candidate retention. Selection should therefore balance broad coverage with attention to parents whose children lie near the pruning threshold, where small score changes can affect retention. Moreover, as the student policy evolves, target prefixes retained at initialization may gradually lose their retention margins, suggesting that initial-reference retention-margin decline can further inform the selection of search-critical states. Second, \emph{what self-teacher knowledge should be transferred at the selected states?} At ground-truth-prefix (GT-prefix) states, whose prefixes match the beginning of the target SID, full KL couples target-probability calibration with conditional preferences among non-target siblings, which are the other legal next-token children of the same parent~\cite{zhao2022dkd}. When the privileged self-teacher assigns high confidence to the observed target, the relative contribution of non-target relational supervision can become small, even though preferences among competing continuations remain informative for search ranking. This motivates separate treatment of target-probability calibration and sibling-relation knowledge.

To address these challenges, we propose \textbf{Search-Aware On-Policy Self-Distillation (SA-OPSD)}. SA-OPSD first collects parent states visited by student beam search and obtains privileged self-teacher distributions conditioned on the training target item title at those same prefixes. For state selection, \textbf{Evolving-Boundary Sampling (EBS)} balances broad coverage with priority for parents whose children lie near the pruning threshold, further incorporating initial-reference retention-margin decline to account for evolving search behavior under a mini-batch distillation budget. For knowledge transfer, \textbf{Confidence-Decoupled Branch Distillation (CDBD)} separates target-probability calibration from the teacher's conditional preferences among non-target siblings, allowing informative branch relations to receive effective supervision even when the self-teacher strongly favors the observed target. Supported non-GT-prefix states receive full KL supervision. The unified objective combines target-SID learning with privileged self-teacher guidance over the student's search states.

We analyze the supervision properties of CDBD and characterize the sampling-weighted objective induced by EBS. Experiments evaluate ranking performance, distillation efficiency, and parameter sensitivity. Additional ablations and search-level diagnostics examine state selection, initial-reference retention-margin decline, sibling-order agreement, and target-prefix survival.

Our main contributions are summarized below.
\begin{itemize}
\setlength{\itemsep}{1pt}
\setlength{\topsep}{2pt}
\setlength{\parsep}{0pt}
    \item \textbf{We reveal a search-decision mismatch between OPSD supervision and competitive beam search decisions.} Rollout comparisons and ablations highlight the value of supervising competing states and prioritizing critical parents. Diagnostics link reference margin decline to pruning risk, while KL analysis reveals weakened sibling supervision under high teacher confidence.

    \item \textbf{We propose SA-OPSD to align privileged supervision with evolving beam search.} EBS prioritizes states using boundary proximity and margin decline, while CDBD preserves competing branch preferences, addressing where and what to distill.

    \item \textbf{We evaluate SA-OPSD across three recommendation benchmarks.} Results show gains over OPD and OPSD baselines on Industrial and Office, while ablations and search diagnostics support both components.
\end{itemize}

\section{Related Work}

\subsection{Generative Recommendation}
Generative recommendation formulates item retrieval as the autoregressive generation of semantic IDs (SIDs) from user interaction histories~\cite{rajput2023tiger,zheng2023lcrec,wang2024letter}. Prior studies improve item tokenization, incorporate collaborative signals into SID construction, and explore the structure of SID hierarchies~\cite{fu2026diger,xie2025lohrec,zheng2026topogr,ju2025grid}. During inference, beam search expands candidates from multiple parent prefixes and selects the highest-scoring continuations across shared beam slots. Consequently, a relevant item can be lost when one of its prefixes is pruned. Recent approaches address this search behavior more explicitly: APAO improves target-prefix survival~\cite{yu2026apao}, BEAR introduces beam-search-aware regularization~\cite{yang2026bear}, and UniSGR studies the connection between SID generation and ranking~\cite{sun2026unisgr}. S-GRec~\cite{jiang2026sgrec} introduces train-time personalåized semantic judging and asymmetric advantage optimization on MiniOneRec-style benchmarks, while V-STAR~\cite{jiang2026vstar} studies value-guided prefix exploration and sibling-relative reinforcement. Our work instead transfers privileged self-teacher knowledge to the competing states encountered during the student's own beam search.

\subsection{Distillation for Recommendation}
Knowledge distillation transfers predictive information to efficient student models~\cite{hinton2015distilling,gou2021survey}. In recommendation, early methods distill item rankings and preference representations~\cite{tang2018ranking,kang2023hetcomp}, while more recent approaches transfer knowledge from language models to recommendation models~\cite{cui2024dllm2rec,chen2025c2kd}. For generative recommendation, SID-MLP distills autoregressive prediction knowledge into lightweight heads~\cite{guo2026sidmlp}, SmartGR uses SID-level and teacher-beam ranking supervision~\cite{zhang2026smartgr}, and Gryphon-v2 distills ranking signals from decoder rollouts~\cite{lipkina2026gryphon}. Student-aware distillation has been explored for recommendation reasoning traces~\cite{shahgir2026scored}. Beyond the supervision source, the form of the distillation objective matters: Decoupled Knowledge Distillation (DKD) separates target-class calibration from non-target distribution matching~\cite{zhao2022dkd}. We apply this distinction to conditional preferences among legal sibling continuations, where relative branch probabilities influence candidate competition.

\subsection{On-Policy Distillation and Self-Distillation}
Conventional KD often supervises target sequences or teacher-generated outputs, including sequence-level distillation~\cite{kim2016sequence}. On-policy distillation (OPD) instead evaluates teacher guidance on student-generated trajectories, as in GKD~\cite{agarwal2023gkd}. Subsequent work explores informative token positions and rollout prefixes~\cite{xu2026tip,zhang2026fastopd}. In recommendation, EvoReason uses on-policy distillation to align latent reasoning with a student's evolving behavior~\cite{zhuang2026evoreason}. Separately, privileged self-distillation and reasoning-based self-distillation have been studied for recommendation~\cite{tang2026psd,jiang2026selfdr}. On-policy self-distillation (OPSD) combines student-generated trajectories with supervision from a same-origin teacher given additional training-time context~\cite{zhao2026opsd}. Building on this paradigm, SA-OPSD queries the privileged self-teacher at student-visited SID prefixes and allocates supervision by their relevance to evolving beam-search behavior.
\begin{figure*}[t]
\centering
\includegraphics[width=0.99\textwidth]{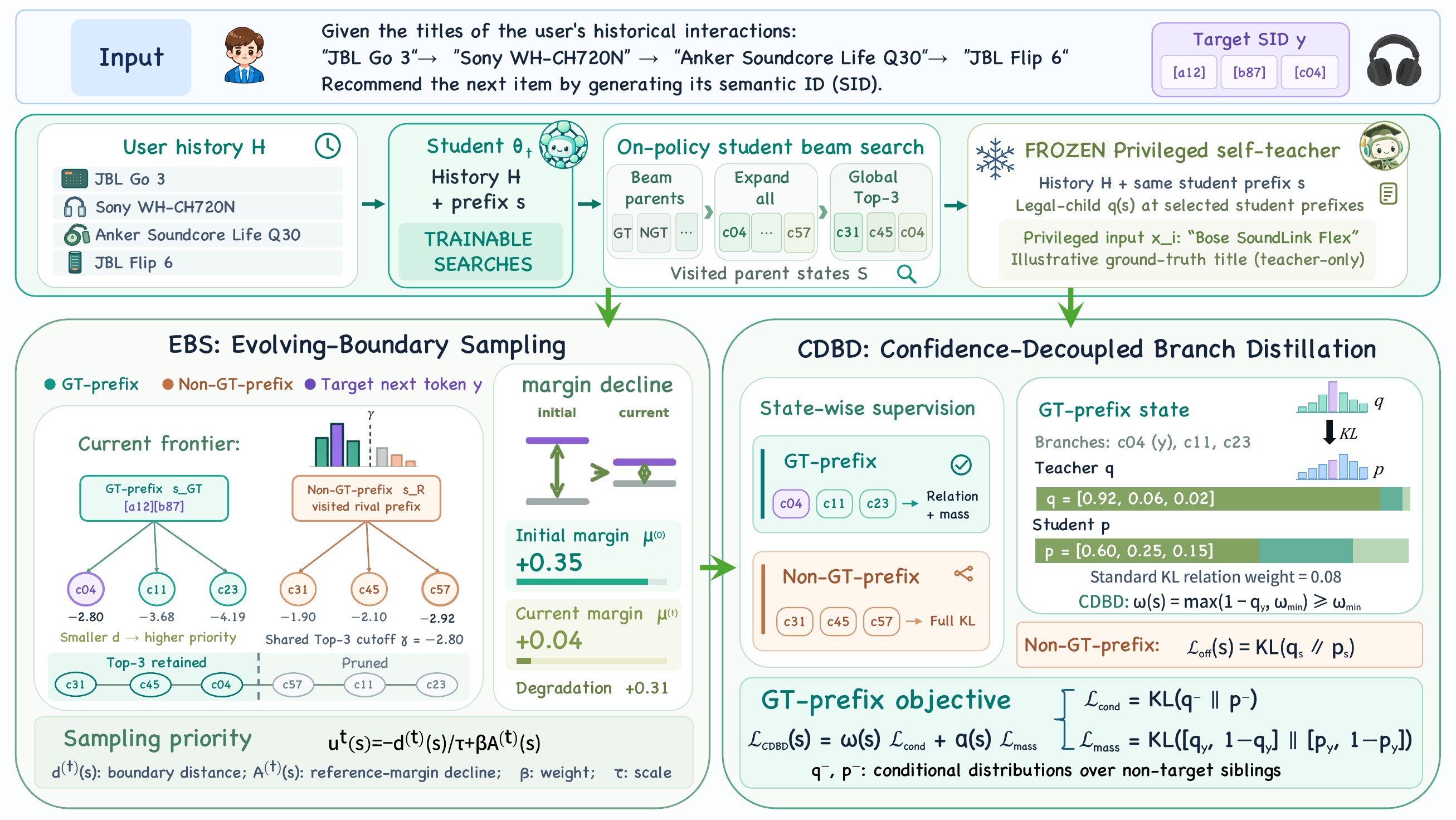}
\caption{Overview of SA-OPSD. Only the student performs beam search. EBS selects parent states using current pruning-boundary proximity and reference-margin decline. A frozen privileged self-teacher receives the same student prefix and the ground-truth item title, and CDBD transfers its conditional branch preferences.}
\Description{The student generates beam-search states. EBS selects parent states, then the privileged self-teacher evaluates legal next-token continuations. CDBD applies relational and target-mass supervision at GT-prefix states and full KL at non-GT-prefix states before the student update.}
\label{fig:overview}
\end{figure*}
\section{Method}
\label{sec:method}
Figure~\ref{fig:overview} presents \method{}, which transfers privileged self-teacher preferences at parent states visited by the student's beam search. EBS allocates a limited supervision budget using pruning-boundary proximity and reference-margin decline, while CDBD preserves conditional preferences among competing siblings. CDBD attenuates target-mass gradients that oppose local hard-label supervision.

\subsection{Search-Aware On-Policy Self-Distillation}
\label{sec:search-states}
\label{sec:search-supervision}
We first formalize the student-generated states used for on-policy supervision. Let $H$ denote a user's interaction history and $y_{1:L}$ the SID of the observed target item. At depth $\ell$, a search state is $s=(H,\hat y_{1:\ell-1})$, where $\hat y_{1:\ell-1}$ is a prefix generated by the student. Under valid-item constraints, $\mathcal C(s)$ denotes its legal next-token children, and $p_{\theta_t}(c\mid s)$ is the student distribution normalized over this set. A candidate prefix $b$ has cumulative log-probability score $z_{\theta_t}(b)$, with $z_{\theta_t}(sc)=z_{\theta_t}(s)+\log p_{\theta_t}(c\mid s)$ for child $c$. At each depth, beam search expands retained prefixes into a frontier $\frontier_\ell$ and keeps its $B$ highest-scoring candidates according to
\begin{equation}
\beam_\ell=\operatorname{TopB}_{b\in\frontier_\ell}z_{\theta_t}(b).
\label{eq:beam-search}
\end{equation}

For interaction $i$, student beam search yields a set $\mathcal S_i$ of visited parent states across decoding depths. A visited state is a \emph{GT-prefix state} when its prefix matches $y_{1:\ell-1}$, in which case $y_\ell$ is the observed next-token label; the remaining visited states are \emph{non-GT-prefix states}. Together, these states represent the target trajectory and its competing search branches.

\paragraph{Privileged self-teacher and supported supervision.}
The student $S_{\theta_t}$ and privileged self-teacher $T_{\bar\theta}$ share the \texttt{Qwen2.5-1.5B-}\allowbreak\texttt{Instruct} backbone and task-adapted initialization. The student conditions on user history $H_i$ and its own generated SID prefix. During training, the frozen self-teacher additionally receives the \emph{complete product title} $x_i=\operatorname{Title}(v_i)$ of the ground-truth item $v_i$. Crucially, both models evaluate the \emph{same student-generated search prefix} $s$: the teacher does not generate a separate search trajectory. At each selected parent, the teacher supplies $q_{i,s}=q_{\bar\theta}(\cdot\mid H_i,x_i,s)$ and the student supplies $p_{\theta_t}(\cdot\mid H_i,s)$, with both distributions normalized over the same legal-child set $\mathcal C(s)$. The target title is unavailable to the student and is never used at inference. Before querying the teacher, we form the structurally eligible set
\begin{equation}
\mathcal E_i=\{s\in\mathcal S_i:|\mathcal C(s)|\ge1\},\qquad M_i=|\mathcal E_i|.
\label{eq:eligible}
\end{equation}
Target SIDs follow the same legal SID tree as constrained decoding, so the observed next token at a visited GT-prefix state belongs to $\mathcal C(s)$. Equation~\eqref{eq:eligible} specifies structural eligibility from the SID constraints and student rollout. EBS allocates query slots over $\mathcal E_i$, with teacher-dependent numerical checks performed after sampling (Sec.~\ref{sec:branch-distillation}).
Let $\theta_0$ denote the unprivileged student at the start of self-distillation. Its initial search behavior serves as EBS's reference, while subsequent mini-batches run beam search only with the current student. Legal SID constraints and cumulative path-scoring conventions are the same across the two student checkpoints.

\subsection{Search-Decision Mismatch and Supervision Challenges}
\label{sec:search-challenges}
\paragraph{The role of competing branches.}
Target-SID cross-entropy supervises the target path but not a non-GT parent $s$, whose children compete for beam slots. The number of its children outranking target $y$ at the final expansion is
\begin{equation}
N_s(y)=\sum_{c\in\mathcal C(s)}
\mathbb I\!\left[z_\theta(s)+\log p_\theta(c\mid s)>z_\theta(y)\right].
\label{eq:competing-threshold}
\end{equation}
A non-GT prefix cannot reach the target SID; distillation there transfers teacher preferences among competing continuations whose cumulative scores affect target retention and rank. SA-OPSD thus supervises GT-prefix and non-GT parents visited by the student (Fig.~\ref{fig:motivation}).

\paragraph{Two challenges of search-state self-distillation.}
Dense branch KD over all visited parents is expensive, so a limited teacher-query budget must prioritize search-critical decisions while preserving interaction coverage. Parents with children near the keep/prune boundary are sensitive to score changes that may alter retention. Yet current boundary proximity cannot capture evolving decisions: initially safe target prefixes may gradually lose retention margins. State selection thus combines current boundary proximity and initial-reference retention-margin decline while maintaining coverage.

At GT-prefix states, SFT directly supervises the observed next token, whereas the privileged self-teacher also provides conditional preferences among non-target siblings. Under full KL, the conditional relation term is weighted by $1-q_y$, so high teacher target confidence weakens its contribution even when sibling ordering remains informative for search. Target-mass calibration introduces a separate issue: when $q_y<p_y$, its gradient opposes local hard-label supervision. These effects motivate decoupled control of target-mass calibration and sibling-relation strength. EBS and CDBD thus address the allocation and content of supervision at student-visited search states, respectively.

\subsection{Evolving-Boundary Sampling}
\label{sec:rbs-budget}
\paragraph{Locating boundary-sensitive decisions.}
We identify sensitive parents using the student's current expansion frontier. For interaction $i$ at depth $\ell$, if $\frontier_{i,\ell}$ contains more than $B$ children, its $B$-th largest cumulative score defines the pruning threshold:
\begin{equation}
\gamma_{i,\ell}=z_{\theta_t}\bigl(b_{i,\ell}^{(B)}\bigr),\quad
b_{i,\ell}^{(B)}=\operatorname{Bth}_{b\in\frontier_{i,\ell}}z_{\theta_t}(b).
\label{eq:pruning-threshold}
\end{equation}
With deterministic tie-breaking, the depth-normalized distance from an expanded parent $s$ to this threshold is
\begin{equation}
d_i^{(t)}(s)=\frac{1}{\ell}\min_{c:\,sc\in\frontier_{i,\ell}}
\bigl|z_{\theta_t}(sc)-\gamma_{i,\ell}\bigr|.
\label{eq:frontier-distance}
\end{equation}
When $|\frontier_{i,\ell}|\le B$, all candidates survive and we set $d_i^{(t)}(s)=+\infty$. Distances use detached scores, with small values identifying parents with children near the keep/prune boundary.

\paragraph{Reference-margin decline.}
Current frontier distance alone does not indicate whether a retention decision has deteriorated over training. For the target prefix $g_{i,\ell}=y_{i,1:\ell}$, let $b_{i,\ell,\neg g}^{(B,t)}$ be the $B$-th highest-scoring \emph{non-target} child in the current frontier. Whenever $g_{i,\ell}$ and at least $B$ non-target children are actually expanded, define its signed retention margin as
\begin{equation}
\mu_{i,\ell}^{(t)}=\frac{z_{\theta_t}(g_{i,\ell})-z_{\theta_t}(b_{i,\ell,\neg g}^{(B,t)})}{\ell}.
\label{eq:retention-margin}
\end{equation}
We obtain the reference margin $\mu_{i,\ell}^{(0)}$ by evaluating the initial student once for each training interaction, using the same beam width, legal SID constraints, score definition, and tie-breaking as the current student. Although the frontier competitor may differ across checkpoints, comparing these margins measures the change in the target prefix's retention advantage relative to initialization. At depths with valid margins in both rollouts, we assign a decline signal to target prefixes with strictly positive initial retention margins:
\begin{equation}
D_{i,\ell}^{(t)}=\mathbb I[\mu_{i,\ell}^{(0)}>0]\,
\bigl[\mu_{i,\ell}^{(0)}-\mu_{i,\ell}^{(t)}\bigr]_+.
\label{eq:reference-based-degradation}
\end{equation}
Zero initial margins, including ties resolved by beam tie-breaking, and undefined margins produce no reference-margin decline signal. If a target first fails the student's top-$B$ retention at depth $\ell$, its prefix is still in the \emph{expanded} frontier, so the event is attributed only to the actually visited parent $g_{i,\ell-1}$ and the parent of its threshold competitor (when supported). For each eligible visited parent, let $A_i^{(t)}(s)$ be the maximum valid $D_{i,\ell}^{(t)}$ attributed to $s$, and zero if none exists. Both reference and current margins are treated as detached selection signals.

\paragraph{Balancing evolving-boundary priority and coverage.}
We mix uniform coverage with search-critical priority. Each eligible parent $s\in\mathcal E_i$ receives
\begin{equation}
\begin{aligned}
u_i^{(t)}(s)&=-d_i^{(t)}(s)/\tau+\beta A_i^{(t)}(s),\\
P_i^{(t)}(s)&=\frac{1-\rho}{M_i}+
\rho\,\frac{\exp[u_i^{(t)}(s)]}{\sum_{v\in\mathcal E_i}\exp[u_i^{(t)}(v)]},
\end{aligned}
\label{eq:trajsample}
\end{equation}
The uniform component has weight $1-\rho$ and the prioritized component has weight $\rho$, with $0\le\rho\le1$. The scale $\tau>0$ controls sensitivity to boundary distance: smaller values concentrate the priority near the boundary. The coefficient $\beta\ge0$ weights reference-margin decline. Equation~\eqref{eq:trajsample} defines the first-draw distribution. Sampling proceeds without replacement from detached weights, renormalized over the remaining parents after each draw. If every priority is $-\infty$, the softmax term is defined as $1/M_i$; if positive weights are exhausted, sampling continues uniformly. Setting $\beta=0$ recovers the boundary-guided selector.

\paragraph{Allocating supervision across interactions.}
Let $C$ be the mini-batch budget of selected-parent teacher evaluations. For mini-batch $\mathcal B$, set $\mathcal B_{\mathrm{KD}}=\{i\in\mathcal B:M_i>0\}$ and $C'=\min(C,\sum_{i\in\mathcal B_{\mathrm{KD}}}M_i)$. When $C'\ge|\mathcal B_{\mathrm{KD}}|$, assign one query slot to each eligible interaction before distributing remaining slots proportional to $M_i-k_i$. Otherwise, sample $C'$ distinct eligible interactions and assign one evaluation to each. Thus, $\sum_i k_i=C'$ and $0\le k_i\le M_i$. Interaction $i$ contributes $k_i$ distinct states $\mathcal A_i\subseteq\mathcal E_i$ under Eq.~\eqref{eq:trajsample}. When the budget permits, each eligible interaction receives at least one selected state. EBS prioritizes boundary-sensitive states within each selected interaction.

\subsection{Confidence-Decoupled Branch Distillation}
\label{sec:branch-distillation}
\paragraph{Teacher preferences among sibling branches.}
At a selected parent $s$, the relative score of two legal children $c_i,c_j\in\mathcal C(s)$ depends on their local next-token probabilities:
\begin{equation}
z_{\theta_t}(sc_i)-z_{\theta_t}(sc_j)
=\log\frac{p_{\theta_t}(c_i\mid s)}{p_{\theta_t}(c_j\mid s)}.
\label{eq:sibling-score-ratio}
\end{equation}
The conditional distribution therefore determines sibling ordering within a parent, whereas competition across parents also depends on their cumulative prefix scores. CDBD transfers privileged self-teacher preferences at the selected student states, providing local supervision whose effects on global beam retention are evaluated empirically in Sec.~\ref{sec:mechanism}. At GT-prefix states, however, the target-child confidence and the relative preferences among non-target siblings constitute distinct forms of supervision.

\paragraph{Separating target mass from sibling relations.}
Omit state subscripts at a GT-prefix parent, and let $y$ be its observed next child. Define target-mass and conditional relation terms by
\begin{align}
\lossmass&=\mathrm{KL}\left([q_y,1-q_y] \,\|\, [p_y,1-p_y]\right),
\label{eq:mass}\\
\losscond&=\mathrm{KL}(q^-\|p^-),\quad
\lossrel=(1-q_y)\losscond,
\label{eq:relation}
\end{align}
where $q_c^-=q_c/(1-q_y)$ and $p_c^-=p_c/(1-p_y)$ for $c\ne y$. The exact KL decomposition~\cite{zhao2022dkd} is
\begin{equation}
\mathrm{KL}(q\|p)=\lossmass+(1-q_y)\losscond=\lossmass+\lossrel.
\label{eq:full-decomposition}
\end{equation}
Full KL scales the conditional non-target relation term by $1-q_y$. In OPSD, a privileged self-teacher may assign high confidence to the observed target, making this relation contribution small even when the ordering among alternatives remains informative. This motivates controlling the relation strength independently of target mass, without changing the teacher's conditional ordering.

\paragraph{Confidence-decoupled supervision.}
Let $\operatorname{sg}(x)$ return $x$ in the forward pass and have zero gradient. CDBD introduces a lower bound on the conditional relation weight:
\begin{equation}
\omega(s)=\max\{1-\operatorname{sg}(q_y),\omega_{\min}\},
\quad \omega_{\min}\ge0.
\label{eq:relation-floor}
\end{equation}
Teacher probabilities use the legal-child softmax defined above. We compute $q^-$ and $p^-$ directly by normalizing the non-target logits with log-sum-exp, which is algebraically equivalent to conditioning the full distribution. The relation term is evaluated when there are at least two non-target legal children and finite, positive target and non-target masses. Otherwise, it is zero while valid target-mass supervision is retained. Since teacher probabilities and $\omega$ are constants during the update, its gradient on a non-target logit is proportional to $\omega(s)(p_c^--q_c^-)$ rather than $(1-q_y)(p_c^--q_c^-)$, while its direct target-logit gradient remains zero. A positive $\omega_{\min}$ thus prevents high teacher target confidence \emph{alone} from suppressing reliable relational supervision.

\paragraph{Conflict-attenuated target-mass calibration.}
For target-mass calibration, CDBD uses a detached gate to control gradients that oppose local hard-label supervision. Let $a$ denote the student's legal-child logits and $\mathcal L_{\mathrm{SFT}}(s)=-\log p_y$ use the same legal-child softmax as distillation. For the frozen self-teacher,
\begin{equation*}
\nabla_a\lossmass=-\frac{p_y-q_y}{1-p_y}\nabla_a\mathcal L_{\mathrm{SFT}}(s),\qquad
\frac{\partial\losscond}{\partial a_y}=0.
\end{equation*}
Target-mass matching opposes local hard-label supervision if $q_y<p_y$. We therefore define the gate as
\begin{equation}
\alpha(s)=
\begin{cases}
1,&\operatorname{sg}(q_y)\ge\operatorname{sg}(p_y),\\
\alpha_{\min},&\operatorname{sg}(q_y)<\operatorname{sg}(p_y),
\end{cases}\quad 0\le\alpha_{\min}\le1.
\label{eq:conflict-gate}
\end{equation}
The resulting GT-prefix objective is
\begin{equation}
\mathcal L_{\mathrm{CDBD}}(s)=\omega(s)\losscond(s)+\alpha(s)\lossmass(s).
\label{eq:cdbd}
\end{equation}
When $\alpha_{\min}<1$, the gate attenuates opposing target-mass gradients, while $\omega(s)$ maintains a confidence-independent lower bound on sibling-relation supervision. Although the relation term has zero direct target-logit gradient, changes to non-target logits can still affect the target probability. The gate therefore does not guarantee compatibility of the full CDBD objective with SFT. For supported GT-prefix states, $\alpha_{\min}=1$ and $\omega_{\min}=0$ recover the local loss $\mathcal L_{\mathrm{CDBD}}(s)=\mathrm{KL}(q_s\|p_s)$. Equation~\eqref{eq:effective-branch} additionally scales the branch loss by $1/|\mathcal C(s)|$. Numerical equivalence to a Full-KL baseline requires the same normalization. Appendix~\ref{app:loss-properties} derives the gradient identities.

\paragraph{Supervision at non-GT-prefix states.}
At supported non-GT-prefix parents, self-teacher supervision uses the full distribution over legal next-token children. Thus, for each selected state,
\begin{equation}
\mathcal L_{\mathrm{branch}}(s)=\frac{1}{|\mathcal C(s)|}
\begin{cases}
\mathcal L_{\mathrm{CDBD}}(s),&\text{GT-prefix},\\
\mathrm{KL}(q_s\|p_s),&\text{non-GT-prefix}.
\end{cases}
\label{eq:effective-branch}
\end{equation}
The conditional relation term vanishes when no meaningful alternative ordering exists (fewer than two non-target children). A singleton legal-child set has zero branch loss. Invalid evaluations contribute zero branch loss. The branch objective is aggregated over the sampled states, including these zero contributions, below.

\subsection{Unified Optimization and Training}
\label{sec:optimization}
\paragraph{Budgeted distillation objective.}
Let $\mathcal I_C=\{i\in\mathcal B:k_i>0\}$ denote interactions receiving KD and write $\ell_s=\mathcal L_{\mathrm{branch}}(s)$. For sampled states $\mathcal A_i$, define
\begin{equation}
\widehat{\mathcal L}^{\mathrm{EBS}}_{\mathrm{KD}}=
\begin{cases}
\dfrac1{|\mathcal I_C|}\displaystyle\sum_{i\in\mathcal I_C}
\dfrac1{k_i}\displaystyle\sum_{s\in\mathcal A_i}\ell_s,
& |\mathcal I_C|>0,\\[2pt]
0,&|\mathcal I_C|=0.
\end{cases}
\label{eq:trajagg}
\end{equation}
The estimator averages selected states within each interaction and then the interactions receiving supervision. Conditional on the current student, reference margins, teacher outputs, and numerical-validity masks, $\mathcal L^{\mathrm{EBS}}_{\mathrm{KD}}=\mathbb E_{\mathbf k,\mathcal A}[\widehat{\mathcal L}^{\mathrm{EBS}}_{\mathrm{KD}}]$ denotes the stop-gradient weighted objective induced by EBS. Sampling probabilities and budget allocations are fixed during each update; gradients flow through the selected-state losses. Appendix~\ref{app:loss-properties} characterizes the inclusion probabilities.

\paragraph{Target-SID learning.}
For interaction $i$, let $s^*_{i,\ell}=(H_i,y_{i,1:\ell-1})$ be the teacher-forced target prefix and $L_i$ its SID length. The target loss averages the per-token cross-entropy over each SID and then over the mini-batch:
\begin{equation}
\mathcal L_{\mathrm{SID}}=\frac{1}{|\mathcal B|}\sum_{i\in\mathcal B}
\frac{1}{L_i}\sum_{\ell=1}^{L_i}\underbrace{-\log p_\theta(y_{i,\ell}\mid s^*_{i,\ell})}_{\mathcal L_{\mathrm{SFT}}(s^*_{i,\ell})}.
\label{eq:sid-loss}
\end{equation}
This loss uses the full target SID independently of whether its prefixes survive beam search. We use the target-SID loss as the primary supervision objective and introduce a single coefficient $\lambda_{\mathrm{KD}}$ to control the contribution of EBS-weighted distillation. The overall training objective is given in Eq.~\eqref{eq:overall}.
\begin{equation}
\boxed{
\mathcal L_{\mathrm{distill}}
=\mathcal L_{\mathrm{SID}}
+\lambda_{\mathrm{KD}}\widehat{\mathcal L}^{\mathrm{EBS}}_{\mathrm{KD}}}.
\label{eq:overall}
\end{equation}
The complete training procedure appears as Algorithm~\ref{alg:framework} in Appendix~\ref{app:training-procedure}. Initial-reference margins are evaluated with the initial student. A single current-student beam rollout in each mini-batch provides candidate states for EBS, which then selects parents for CDBD supervision. Candidate and state selection is detached, while gradients flow through the branch loss at selected parents. The deployed student uses the same constrained SID beam decoder without self-teacher access or additional inference stages.

\begin{table}[!t]
\centering
\caption{Statistics of the three datasets.}
\label{tab:dataset-statistics}
\bordtablestyle
\fontsize{8.5}{10.2}\selectfont
\renewcommand{\arraystretch}{1.12}
\setlength{\tabcolsep}{3pt}
\begin{tabularx}{\columnwidth}{@{}>{\hsize=1.33\hsize\raggedright\arraybackslash}X>{\hsize=.89\hsize\raggedleft\arraybackslash}X>{\hsize=.89\hsize\raggedleft\arraybackslash}X>{\hsize=.89\hsize\raggedleft\arraybackslash}X@{}}
\toprule
\rowcolor{BordHeader}
\bordhead{Statistic} & \bordhead{Industrial} & \bordhead{Office} & \bordhead{Instruments} \\
\midrule
\#Users & 7,694 & 8,328 & 7,537 \\
\#Items & 3,685 & 3,459 & 3,399 \\
\#Interactions & 50,354 & 54,226 & 57,482 \\
\hline 
Train samples & 36,259 & 38,924 & 39,956 \\
Valid samples & 4,532 & 4,866 & 4,994 \\
Test samples & 4,533 & 4,866 & 4,995 \\
\bottomrule
\end{tabularx}
\end{table}

\begin{table*}[!t]
\centering
\caption{Retrieval performance on Industrial, Office, and Instruments (HR/NDCG, higher is better). Best and second-best values in each metric are bold and underlined.}
\label{tab:overall}
\bordtablestyle
\fontsize{7.6}{8.9}\selectfont
\renewcommand{\arraystretch}{1.13}
\setlength{\tabcolsep}{2pt}
\begin{tabular*}{\textwidth}{@{\extracolsep{\fill}}l*{12}{r}@{}}
\toprule
\rowcolor{BordHeader} \textbf{Method} & \multicolumn{4}{c}{\bordhead{Industrial}} & \multicolumn{4}{c}{\bordhead{Office}} & \multicolumn{4}{c}{\bordhead{Instruments}} \\
\cmidrule(lr){2-5}\cmidrule(lr){6-9}\cmidrule(lr){10-13}
 & HR@5 & NDCG@5 & HR@10 & NDCG@10 & HR@5 & NDCG@5 & HR@10 & NDCG@10 & HR@5 & NDCG@5 & HR@10 & NDCG@10 \\
\midrule
\rowcolor{BordGroup}\multicolumn{13}{l}{\textcolor{BordInk}{\textit{Sequential recommenders}}} \\
\quad GRU4Rec & 0.0774 & 0.0598 & 0.0999 & 0.0669 & 0.0789 & 0.0595 & 0.1019 & 0.0669 & 0.0661 & 0.0538 & 0.0871 & 0.0605 \\
\quad Caser & 0.0717 & 0.0555 & 0.0942 & 0.0628 & 0.0865 & 0.0664 & 0.1093 & 0.0737 & 0.0628 & 0.0510 & 0.0845 & 0.0580 \\
\quad SASRec & 0.0909 & 0.0748 & 0.1088 & 0.0806 & 0.0949 & 0.0805 & 0.1120 & 0.0858 & 0.0793 & 0.0675 & 0.0971 & 0.0733 \\
\quad BERT4Rec & 0.0869 & 0.0709 & 0.1110 & 0.0784 & 0.0912 & 0.0753 & 0.1173 & 0.0836 & 0.0759 & 0.0652 & 0.0980 & 0.0725 \\
\addlinespace[2pt]
\rowcolor{BordGroup}\multicolumn{13}{l}{\textcolor{BordInk}{\textit{Generative recommenders}}} \\
\quad HSTU & 0.1037 & 0.0918 & 0.1163 & 0.0958 & 0.1252 & 0.1079 & 0.1400 & 0.1126 & 0.0889 & 0.0820 & 0.1027 & 0.0863 \\
\quad TIGER & 0.1010 & 0.0807 & 0.1321 & 0.0908 & 0.1163 & 0.0960 & 0.1408 & 0.1030 & 0.0888 & 0.0742 & 0.1188 & 0.0841 \\
\quad LC-Rec & 0.1057 & 0.0862 & 0.1332 & 0.0952 & 0.1048 & 0.0859 & 0.1237 & 0.0920 & 0.0919 & 0.0788 & 0.1174 & 0.0868 \\
\quad LETTER & 0.1238 & 0.1028 & 0.1491 & 0.1108 & 0.1268 & 0.1064 & 0.1552 & 0.1153 & 0.1079 & 0.0943 & 0.1329 & 0.1025 \\
\quad ActionPiece & 0.1220 & 0.1015 & 0.1487 & 0.1099 & 0.1297 & 0.1077 & 0.1564 & 0.1162 & 0.1076 & 0.0937 & 0.1314 & 0.1015 \\
\quad LOHRec & 0.1275 & 0.1056 & 0.1546 & 0.1142 & 0.1344 & 0.1125 & 0.1617 & 0.1212 & 0.1135 & 0.0988 & 0.1413 & 0.1077 \\
\addlinespace[2pt]
\rowcolor{BordGroup}\multicolumn{13}{l}{\textcolor{BordInk}{\textit{LLM-based recommenders}}} \\
\quad BIGRec & 0.1092 & 0.0907 & 0.1370 & 0.0997 & 0.1204 & 0.1017 & 0.1434 & 0.1091 & 0.0961 & 0.0831 & 0.1233 & 0.0918 \\
\quad S-DPO & 0.1238 & 0.0991 & 0.1524 & 0.1082 & 0.1356 & 0.1110 & 0.1587 & 0.1188 & 0.1079 & 0.0908 & 0.1364 & 0.1000 \\
\quad MiniOneRec & 0.1321 & 0.1084 & 0.1586 & 0.1167 & 0.1420 & 0.1172 & 0.1634 & 0.1242 & 0.1157 & 0.1005 & 0.1424 & 0.1092 \\
\quad S-GRec & 0.1359 & 0.1113 & 0.1632 & 0.1202 & 0.1496 & 0.1246 & 0.1689 & 0.1308 & 0.1165 & 0.0995 & 0.1425 & 0.1078 \\
\addlinespace[2pt]
\rowcolor{BordGroup}\multicolumn{13}{l}{\textcolor{BordInk}{\textit{Search, optimization, and distillation}}} \\
\quad APAO & 0.1328 & 0.1098 & \second{0.1652} & 0.1202 & 0.1425 & 0.1173 & 0.1710 & 0.1262 & 0.1171 & 0.1010 & 0.1491 & 0.1115 \\
\quad SID-MLP & 0.1240 & 0.1021 & 0.1520 & 0.1106 & 0.1299 & 0.1073 & 0.1580 & 0.1165 & 0.1067 & 0.0924 & 0.1326 & 0.1003 \\
\quad SmartGR & \second{0.1368} & \second{0.1142} & 0.1648 & \second{0.1229} & 0.1459 & 0.1216 & 0.1697 & 0.1295 & \second{0.1230} & \second{0.1075} & \second{0.1496} & \second{0.1158} \\
\quad V-STAR & 0.1361 & 0.1128 & 0.1641 & 0.1217 & \second{0.1500} & \second{0.1260} & \second{0.1746} & \best{0.1340} & 0.1203 & 0.1042 & 0.1478 & 0.1130 \\
\textbf{SA-OPSD (Ours)} & \best{0.1410} & \best{0.1173} & \best{0.1690} & \best{0.1263} & \best{0.1523} & \best{0.1262} & \best{0.1747} & \second{0.1334} & \best{0.1267} & \best{0.1100} & \best{0.1526} & \best{0.1192} \\
\bottomrule
\end{tabular*}
\end{table*}

\begin{table}[!ht]
\centering
\caption{NDCG@10 under trajectory-based and beam-search-based distillation.}
\label{tab:rollout-strategies}
\bordtablestyle
\fontsize{8}{9.5}\selectfont
\setlength{\tabcolsep}{3pt}
\begin{tabularx}{\columnwidth}{
@{}l
>{\centering\arraybackslash}X
>{\centering\arraybackslash}X
>{\centering\arraybackslash}X@{}
}
\toprule
\rowcolor{BordHeader}
\bordhead{Strategy} &
\bordhead{Industrial} &
\bordhead{Office} &
\bordhead{Instruments} \\
\midrule
Single Rollout & 0.1168 & 0.1233 & 0.1093 \\
Multi Rollout & 0.1198 & 0.1266 & 0.1122 \\
Dense Beam (Full KL) &
\second{0.1251} & \second{0.1277} & \second{0.1149} \\
\rowcolor{BordTint}
\bordname{SA-OPSD (Ours)} &
\best{0.1263} & \best{0.1334} & \best{0.1192} \\
\bottomrule
\end{tabularx}
\end{table}

\section{Experiments}
\label{sec:experiments}
We systematically evaluate SA-OPSD on three recommendation benchmarks in terms of retrieval accuracy, component contributions, search behavior, and training efficiency. Parameter sensitivity is further reported in Appendix E.

\subsection{Experimental Settings}
\label{sec:experimental-settings}
\paragraph{Datasets and preprocessing.}
We evaluate the Amazon Reviews 2018 \emph{Industrial and Scientific} (Industrial), \emph{Office Products} (Office), and \emph{Musical Instruments} (Instruments) datasets (Table~\ref{tab:dataset-statistics}). Following MiniOneRec~\cite{kong2025minionerec}, we filter users and items with fewer than five interactions, retain the ten most recent history items, and use chronological 8:1:1 splits.

\paragraph{Evaluation protocol.}
For each example, the model predicts a target SID from the user's interaction history. Matched student variants share the constrained beam decoder. We report HR and NDCG at $K=5,10$, using NDCG@10 as the primary metric.

\paragraph{Model backbone and training protocol.}
The student and privileged self-teacher use \texttt{Qwen2.5-1.5B-}\allowbreak\texttt{Instruct}. Matched variants share student initialization, data, target-SID objective, legal-child space, and decoding. Only the privileged self-teacher receives the target item title during training. We fix the target-SID loss coefficient to 1 and use $\lambda_{\mathrm{KD}}$ as the single relative weight for distillation in Eq.~\eqref{eq:overall}. EBS evaluates initial-student margins once under matched beam-search settings and uses declines only where target-prefix frontiers are actually expanded. State-selection ablations supervise 460 parent states under coverage-first allocation, whereas the CDBD ablation changes only branch supervision. The literature baselines are detailed in Appendix~A.

\subsection{Overall Performance}
\label{sec:overall}
Table~\ref{tab:overall} compares recommendation accuracy across the three datasets. SA-OPSD leads the listed methods on all four metrics for Industrial and Instruments. Its NDCG@10 exceeds SmartGR by 0.0034 on both Industrial and Instruments. On Office, SA-OPSD leads in HR@5, NDCG@5, and HR@10, while its NDCG@10 is slightly below V-STAR (0.1334 versus 0.1340). The following ablations examine the contributions of EBS and CDBD.

\subsection{Ablation Study}
\label{sec:ablation}
\begin{figure}[t]
\centering
\includegraphics[width=0.98\columnwidth]{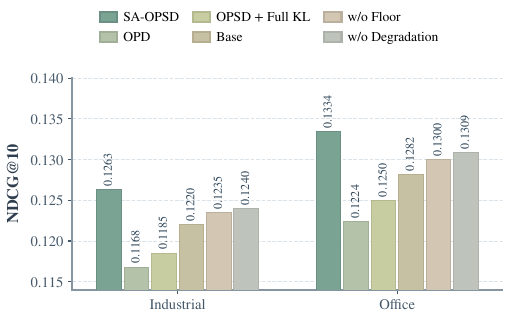}
\caption{Industrial and Office NDCG@10 ablations. OPD and OPSD + Full KL are reference controls. Base, w/o Floor, w/o Degradation, and SA-OPSD isolate reference-margin decline priority ($\beta>0$) and the non-target relation floor ($\omega_{\min}>0$). All four retain the target-mass gate.}
\Description{Grouped bars compare Industrial and Office on a shared NDCG at ten axis. Industrial NDCG at ten: OPD 0.1168, OPSD plus Full KL 0.1185, Base 0.1220, without Floor 0.1235, without Degradation 0.1240, SA-OPSD 0.1263. Office NDCG at ten: OPD 0.1224, OPSD plus Full KL 0.1250, Base 0.1282, without Floor 0.1300, without Degradation 0.1309, SA-OPSD 0.1334.}
\label{fig:ablation-main}
\end{figure}

\paragraph{Rollout strategies and search alignment.}
To examine the search-decision mismatch, Table~\ref{tab:rollout-strategies} compares broader trajectory sampling with supervision at competing student beam-search states. Relative to single-rollout, multi-rollout raises NDCG@10 by 0.0030, 0.0033, and 0.0029 on Industrial, Office, and Instruments, respectively, indicating benefits from broader trajectory coverage. Dense Beam (Full KL) provides further gains of 0.0053, 0.0011, and 0.0027 over multi-rollout, consistent with the value of exposing prefixes that directly compete for shared beam slots. However, observing more beam states is not the whole story. SA-OPSD further improves over Dense Beam (Full KL) by 0.0012, 0.0057, and 0.0043. The multi-to-dense gain is larger on Industrial, whereas the subsequent SA-OPSD gain is larger on Office and Instruments. This contrast suggests that the benefit of search alignment varies by dataset and that state prioritization and branch supervision can complement beam-state coverage. As Table~\ref{tab:rollout-strategies} varies these choices jointly, we use the controlled ablations in Table~\ref{tab:opsd-reference-selection} and Fig.~\ref{fig:ablation-main} to examine them more directly.

\paragraph{Component ablations.}
We compare OPD and OPSD + Full KL as reference controls and use a $2\times2$ ablation to measure the contributions of reference-margin decline priority and the relation floor. OPD uses student-generated prefixes and Full KL without privileged target-title input; OPSD + Full KL uses the privileged self-teacher and Full KL at sampled student-beam parents. All four component configurations use privileged OPSD, current student-beam frontiers, a shared coverage-first 460-state KD budget, and the same target-mass gate $\alpha(s)$ at GT-prefix states. They differ only in the reference-margin decline coefficient $\beta\in\{0,0.4\}$ and relation floor $\omega_{\min}\in\{0,0.15\}$: Base $(0,0)$, w/o Floor $(0.4,0)$, w/o Degradation $(0,0.15)$, and SA-OPSD $(0.4,0.15)$.

\paragraph{Isolating reference-margin decline priority and the relation floor.}
When the conditional-relation floor is disabled ($\omega_{\min}=0$), CDBD specializes to the \emph{gated KL} loss at a supported GT-prefix state:
\begin{equation}
\mathcal L_{\mathrm{gated}}(s)=\alpha(s)\lossmass(s)
     +(1-q_y)\losscond(s).
\label{eq:gated-kl-ablation}
\end{equation}
Base and w/o Floor use this gated KL objective, while w/o Degradation and SA-OPSD add the relation floor. Boundary-based sampling, the target-mass gate $\alpha(s)$, and supported-child checks are common to all four configurations.

\paragraph{Component effects and conclusions.}
Both reference-margin decline priority and the relation floor improve NDCG@10 over Base on both datasets under the same supervision budget. The former prioritizes states with declining retention margins, while the latter strengthens supervision of non-target sibling relations. Combining them yields gains of 0.0043 on Industrial and 0.0052 on Office, outperforming either single-component variant. SA-OPSD surpasses OPSD + Full KL by 0.0078 and 0.0084, respectively. With current-boundary priority and target-mass gating shared across all four configurations, these results support the incremental benefits of reference-margin decline priority and the relation floor.

\subsection{Mechanism Analysis}
\label{sec:mechanism}
\label{sec:search-state-evidence}

\begin{table}[t]
\centering
\caption{State selection on Industrial. Regression (\%) is the fraction of initially correct Top-10 cases lost after training.}
\label{tab:opsd-reference-selection}
\bordtablestyle
\fontsize{7.55}{9.1}\selectfont
\renewcommand{\arraystretch}{1.16}
\setlength{\tabcolsep}{2.3pt}
\begin{tabularx}{\columnwidth}{@{}>{\raggedright\arraybackslash}Xrr@{}}
\toprule
\rowcolor{BordHeader}\textbf{Sampling Strategy} & \textbf{NDCG@10} & \textbf{Regression (\%)} \\
\midrule
Uniform sampling & 0.1216 & 16.1\% \\
Current boundary only & 0.1240 & 14.5\% \\
Boundary + shuffled reference declines & \second{0.1248} & \second{14.4\%} \\
\rowcolor{BordTint}\bordname{EBS (Ours)} & \textbf{0.1263} & \textbf{10.7\%} \\
\bottomrule
\end{tabularx}
\end{table}

\paragraph{EBS: reference-aware state selection.}
We next examine whether visited beam-search states warrant equal supervision priority. Under the same CDBD objective and 460-state KD budget, current-boundary sampling improves NDCG@10 over uniform sampling from 0.1216 to 0.1240, and EBS further reaches 0.1263 (Table~\ref{tab:opsd-reference-selection}). Within similar current-margin ranges, larger reference-margin declines are associated with higher subsequent pruning risk (Appendix~\ref{app:reference-risk}, Fig.~\ref{fig:reference-risk}). EBS also reduces regression of initially correct Top-10 predictions from 14.5\% under current-boundary sampling to 10.7\%. Shuffling decline values weakens these gains, supporting the value of state-specific reference information for supervision allocation.

\begin{figure}[t]
\centering
\includegraphics[width=0.96\columnwidth]{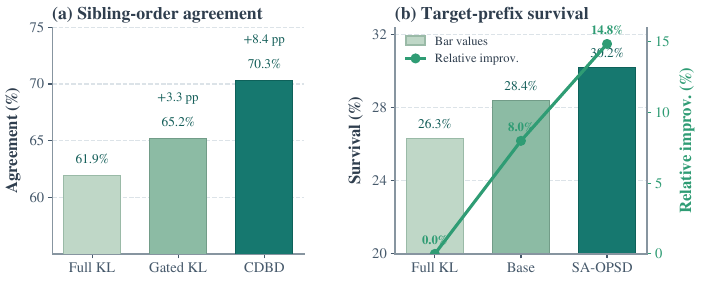}
\caption{Industrial search diagnostics. (a) Sibling-order agreement: bars show percentages, and the +3.3 pp and +8.4 pp annotations denote absolute percentage-point gains over Full KL. (b) Depth-four GT-prefix survival: bars show percentages, while the line shows relative gains (\%) over Full KL.}
\Description{The left panel shows sibling-order agreement of 61.9, 65.2, and 70.3 percent for Full KL, Gated KL, and CDBD. The right panel shows depth-four target-prefix survival of 26.3, 28.4, and 30.2 percent for Full KL, Base, and SA-OPSD, plus a relative-improvement line of 0.0, 8.0, and 14.8 percent.}
\label{fig:sibling-analysis}
\end{figure}

\paragraph{CDBD: sibling ordering.}
At GT-prefix states with high teacher confidence, full KL assigns limited weight to non-target sibling relations, weakening the transfer of competing branch preferences. To examine this effect, we compare Full KL, gated KL, and CDBD on the same student-beam states. Target-mass gating improves agreement with teacher sibling rankings, while the relation floor provides further gains (Fig.~\ref{fig:sibling-analysis}(a)). This indicates that preserving non-target relations complements target-mass calibration and improves the transfer of relative preferences among continuations.

\paragraph{Joint search behavior.}
We measure cumulative GT-prefix survival at each SID depth using beam search with the same decoding width. On Industrial with beam width 50, depth-four survival increases from 26.3\% under student-beam OPSD + Full KL to 28.4\% with Base and 30.2\% under full SA-OPSD (Fig.~\ref{fig:sibling-analysis}(b)). Fewer target paths are thus pruned before reaching the deeper SID positions, consistent with the NDCG@10 gains in Fig.~\ref{fig:ablation-main}.

\subsection{Efficiency Analysis}
\label{sec:efficiency}
\begin{figure}[t]
\centering
\includegraphics[width=0.99\columnwidth]{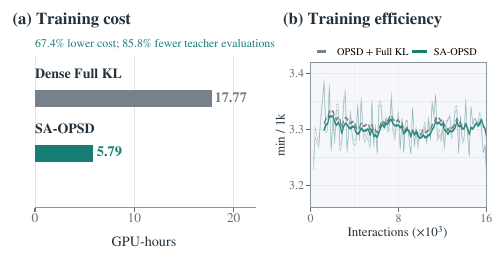}
\caption{Training efficiency. (a) Training Cost Compared with Dense Beam (Full KL). (b) Time per 1k training interactions for OPSD + Full KL and SA-OPSD on Industrial.}
\Description{Panel a shows the Dense Beam (Full KL) configuration at 17.77 GPU-hours with 470,800 teacher-state evaluations, and SA-OPSD at 5.79 GPU-hours with 66,742 evaluations, a 67.4 percent training cost reduction. Panel b shows the per-one-thousand-interaction timing curves near 3.3 minutes for OPSD plus Full KL and SA-OPSD.}
\label{fig:training-comparison}
\end{figure}

\paragraph{Training cost and per-interaction runtime.}
SA-OPSD limits teacher evaluations through a selected-parent budget rather than querying every visited parent. Within this budget, EBS prioritizes states using current boundary proximity and initial-reference margin decline. As shown in Fig.~\ref{fig:training-comparison}(a), this budgeted strategy reduces teacher-state evaluations by 85.8\% and training cost by 67.4\% compared with Dense Beam (Full KL). The smaller reduction in GPU-hours reflects the remaining costs of student beam search and optimization. Meanwhile, Fig.~\ref{fig:training-comparison}(b) shows comparable per-interaction runtime to OPSD + Full KL, indicating that EBS and CDBD introduce little additional computational overhead. These results show lower training cost than Dense Beam (Full KL) and comparable per-interaction runtime to OPSD + Full KL, while Fig.~\ref{fig:ablation-main} separately reports the retrieval improvements. Both components operate only during training, leaving inference unchanged.

\section{Conclusion}
We studied how on-policy self-distillation can reflect the competitive search behavior of generative recommenders. We proposed SA-OPSD, which distills privileged self-teacher knowledge at student-visited beam-search states. Evolving-Boundary Sampling directs supervision toward boundary-sensitive states using current pruning proximity and reference-margin decline, while Confidence-Decoupled Branch Distillation preserves sibling preferences under high teacher target confidence. Experiments across three recommendation benchmarks demonstrate strong retrieval performance. Ablations and search diagnostics support improvements in state selection, sibling ordering, and target-prefix survival. Budgeted supervision also reduces training cost relative to Dense Beam (Full KL) without modifying beam-search inference. These findings highlight the value of search-aligned on-policy distillation.
\clearpage
\newpage

\bibliographystyle{ACM-Reference-Format}
\bibliography{references}
\appendix
\nobalance
\renewcommand{\dbltopfraction}{0.80}
\section{Baselines}
Table~\ref{tab:overall} compares SA-OPSD with results from sequential recommenders (GRU4Rec~\cite{hidasi2016gru4rec}, Caser~\cite{tang2018caser}, SASRec~\cite{kang2018sasrec}, and BERT4Rec~\cite{sun2019bert4rec}), generative recommenders (HSTU~\cite{zhai2024hstu}, TIGER, LC-Rec, LETTER, ActionPiece~\cite{hou2025actionpiece}, and LOHRec), LLM-based recommenders (BIGRec~\cite{bao2025bigrec}, S-DPO~\cite{chen2024sdpo}, MiniOneRec, and S-GRec~\cite{jiang2026sgrec}), and search/optimization/distillation approaches (APAO, SID-MLP, SmartGR, and V-STAR~\cite{jiang2026vstar}). S-GRec reports results with MiniOneRec preprocessing scripts and split statistics, and V-STAR follows MiniOneRec preprocessing. The reported baselines follow their respective published evaluation protocols. Matched student-beam OPSD + Full KL and OPD controls are evaluated alongside the component ablations in Sec.~\ref{sec:ablation} and Fig.~\ref{fig:ablation-main}.

\section{Training Procedure}
\label{app:training-procedure}
Algorithm~\ref{alg:framework} details the one-time reference-margin evaluation and the budgeted on-policy self-distillation updates. All state-selection and sampling decisions are detached from gradient computation.

\begin{algorithm}[H]
\caption{Training procedure of \method{}}
\label{alg:framework}
\footnotesize
\begin{algorithmic}[1]
\Require Training data $\mathcal D=\{(H_i,y_i,x_i)\}$ with target titles $x_i$; initial student $S_{\theta_0}$; privileged self-teacher $T_{\bar\theta}$.
\Statex \hspace{\algorithmicindent} Beam width $B$, budget $C$; $\rho,\tau,\beta,\alpha_{\min},\omega_{\min},\lambda_{\mathrm{KD}}$.
\Statex \textit{Reference search decisions (initial student)}
\For{each $(H_i,y_i,x_i)\in\mathcal D$}
  \State Run constrained beam search with $S_{\theta_0}$ and width $B$.
  \State Record valid target-prefix margins $\mu_{i,\ell}^{(0)}$ via Eq.~\eqref{eq:retention-margin}.
\EndFor
\Statex \textit{On-policy self-distillation}
\For{mini-batch $\mathcal B\subset\mathcal D$ at update step $t$}
  \For{each $(H_i,y_i,x_i)\in\mathcal B$}
    \State Run current student beam search without selection gradients.
    \State Record expanded frontiers, retained beams, and visited parents.
    \State Find valid $\mu_{i,\ell}^{(t)}$ and degradation $D_{i,\ell}^{(t)}$.
    \State Attribute valid margin declines to visited GT and boundary parents.
    \State Exclude unreachable target depths after the first pruning.
    \State Form structural candidates $\mathcal E_i$ by Eq.~\eqref{eq:eligible}; set $M_i\gets|\mathcal E_i|$.
    \State Compute $d_i^{(t)}(s)$ and $A_i^{(t)}(s)$ for $s\in\mathcal E_i$.
  \EndFor
  \State Set $C'\gets\min\{C,\sum_i M_i\}$; allocate $k_i$ by coverage-first budgeting.
  \For{each $i$ with $k_i>0$}
    \State Sample $\mathcal A_i\subseteq\mathcal E_i$ without replacement using Eq.~\eqref{eq:trajsample}.
    \For{each $s\in\mathcal A_i$}
      \State Query $T_{\bar\theta}$ on $(H_i,x_i,s)$ with the same student prefix $s$.
      \State Set $\ell_s\gets0$; count this query against its allocated slot.
      \If{legal-child logits and normalization are numerically valid}
        \State Normalize $q_s$ and $p_s$ over $\mathcal C(s)$.
        \If{$s$ is a GT-prefix state}
          \State Disable the relation term if its support check fails.
          \State $\ell_s\gets\mathcal L_{\mathrm{CDBD}}(s)/|\mathcal C(s)|$.
        \Else
          \State $\ell_s\gets\mathrm{KL}(q_s\|p_s)/|\mathcal C(s)|$.
        \EndIf
      \EndIf
    \EndFor
  \EndFor
  \State Aggregate selected branch losses by Eq.~\eqref{eq:trajagg}.
  \State Compute target-SID loss by Eq.~\eqref{eq:sid-loss}; update $\theta_t$ by Eq.~\eqref{eq:overall}.
\EndFor
\State \Return trained student $S_\theta$ (unchanged constrained beam decoder).
\end{algorithmic}
\end{algorithm}

\section{Reference-Margin Decline and Pruning Risk}
\label{app:reference-risk}
\begin{figure}[H]
\centering
\includegraphics[width=0.94\columnwidth]{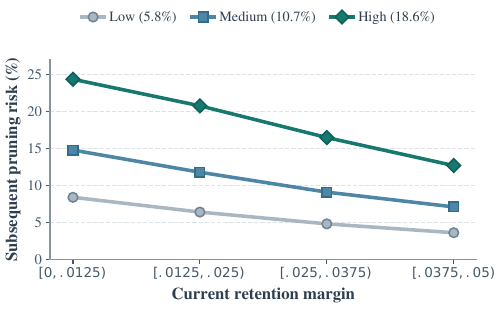}
\caption{Industrial: subsequent first-pruning risk across current retention-margin bins, stratified by low, medium, and high reference-margin decline. Legend values are simple, unweighted averages over the four bins.}
\Description{Three lines display subsequent first-pruning risk across four current-margin bins, from zero to 0.05. Risks decrease with increasing current margin in every reference-margin decline group. High degradation has the greatest risk at each margin bin, followed by medium and low degradation.}
\label{fig:reference-risk}
\end{figure}

For each target SID, we define its first-pruning depth as the earliest depth $\ell$ at which its prefix enters the expanded frontier $\mathcal F_{i,\ell}$ but is not retained in the beam $\mathcal B_{i,\ell}$. Each such event is counted once and attributed to the last visited GT-prefix parent. Initial and current margins are computed with identical beam width, SID depth, decoding constraints, and path-score normalization. Cases are stratified by current retention margin and reference-margin decline. The four margin bins are $[0,0.0125)$, $[0.0125,0.025)$, $[0.025,0.0375)$, and $[0.0375,0.05)$. Figure~\ref{fig:reference-risk} reports the proportion of tracked cases with a subsequent first-pruning event within each margin-bin and ordered reference-decline group.

Within each current-margin bin, larger reference-margin decline is associated with higher subsequent pruning risk. In the first bin, the Low, Medium, and High groups have risks of 8.4\%, 14.8\%, and 24.4\%, respectively. Their unweighted averages across the four bins are 5.8\%, 10.7\%, and 18.6\%. This ordering persists across all four bins, adding a reference-margin perspective to the state-selection comparisons in Table~\ref{tab:opsd-reference-selection}.

\section{Properties of EBS and CDBD}
\label{app:loss-properties}
The sampling analysis conditions on the recorded beam-search frontiers and uses detached selection weights. The gradient analysis concerns a single GT-prefix state in the legal-child logit space, with SFT and distillation sharing the same normalization and temperature. Updates of the shared model parameters combine gradients across states.

\paragraph{Evolving-boundary-weighted sampling property.}
Condition on the current student $\theta_t$, the initial reference $\theta_0$ and corresponding initial-reference margins, the eligible states and frontier scores, the privileged self-teacher outputs and numerical-validity masks at those states, and budget allocation $\mathbf k$. The EBS probability $P_i^{(t)}(s)$ from Eq.~\eqref{eq:trajsample} depends on both detached current frontier distance and the reference-margin decline score. Let $\pi_{i,s}^{(t)}(\mathbf k)$ be the \emph{actual} inclusion probability of state $s$ after $k_i$ sequential draws without replacement. In general $\pi_{i,s}^{(t)}(\mathbf k)\ne k_iP_i^{(t)}(s)$. Since the losses $\ell_s$ are deterministic under these conditions, linearity of expectation gives
\begin{equation}
\mathbb E\!\left[\frac1{k_i}\sum_{s\in\mathcal A_i}\ell_s\,\middle|\,\mathbf k,\theta_t,\theta_0\right]
=\frac1{k_i}\sum_{s\in\mathcal E_i}\pi_{i,s}^{(t)}(\mathbf k)\ell_s,\quad k_i>0.
\label{eq:bg-inclusion}
\end{equation}
Consequently, the budget-and-state expectation of Eq.~\eqref{eq:trajagg}, conditional on $\theta_t,\theta_0$, is
\begin{equation}
\begin{aligned}
\mathcal L_{\mathrm{KD}}^{\mathrm{EBS}}(\theta_t\mid\theta_0;C,\rho,\tau,\beta)
:=\mathbb E_{\mathbf k}\!\biggl[
&\frac{\mathbb I[|\mathcal I_C|>0]}{\max(1,|\mathcal I_C|)}\\[-1pt]
&\sum_{i\in\mathcal I_C}\frac1{k_i}
\sum_{s\in\mathcal E_i}\pi_{i,s}^{(t)}(\mathbf k)\ell_s
\biggr].
\end{aligned}
\label{eq:bg-objective}
\end{equation}
The objective is zero when $C'=0$. Its inclusion probabilities specify stop-gradient weights at the current student checkpoint, and each update differentiates the sampled losses with these weights fixed. Setting $\beta=0$ yields the boundary-weighted sampler.

For fixed frontier distances, $\rho>0$, and $\beta>0$, its softmax priority is nondecreasing in the state's reference-margin decline by construction. The first-draw probabilities follow Eq.~\eqref{eq:trajsample}, while the multi-draw objective uses the inclusion probabilities in Eq.~\eqref{eq:bg-inclusion}. Table~\ref{tab:opsd-reference-selection} evaluates the state-selection choices empirically.

\paragraph{Full-KL expansion.}
The Full-KL decomposition follows by substituting $q_c=(1-q_y)q_c^-$ and $p_c=(1-p_y)p_c^-$ for each $c\neq y$ at a GT-prefix state and collecting terms.
\begin{equation}
\begin{aligned}
\sum_{c\neq y}q_c\log\frac{q_c}{p_c}
={}&(1-q_y)\log\frac{1-q_y}{1-p_y}\\
&+(1-q_y)\mathrm{KL}(q^-\|p^-).
\end{aligned}
\end{equation}
Adding the target-child term $q_y\log(q_y/p_y)$ gives the target/non-target decomposition~\cite{zhao2022dkd}.
\begin{equation}
\lossfull=\mathrm{KL}(q\|p)=\lossmass+\lossrel.
\label{eq:decomp}
\end{equation}
CDBD replaces the coupled KL weights with $\alpha(s)$ for target mass and $\omega(s)$ for conditional sibling relations.

The mass term captures target-versus-rest probability, whereas the conditional relation term captures preferences among non-target children. With $m\ge2$ legal children, target mass has one degree of freedom, and the non-target conditional distribution has $m-2$. Hence, the relation term vanishes when there are only two legal children. For a singleton legal-child set, both distributions equal $[1]$, and the branch loss is defined directly as zero. Zero teacher mass follows the standard KL limit convention. Division by $m$ preserves the decomposition under legal-child normalization.

\paragraph{Local score interpretation.}
Subtracting the cumulative log scores of two children of the same parent gives Eq.~\eqref{eq:sibling-score-ratio}. For non-target siblings, conditioning cancels the shared factor $1-p_y$, yielding $p(c_i\mid s)/p(c_j\mid s)=p_i^-/p_j^-$. Matching conditional sibling probabilities therefore transfers relative scores within a parent. The detached gate $\alpha(s)$ adjusts target-mass calibration, while $\omega(s)$ preserves a lower bound on conditional sibling-relation strength despite high teacher target confidence.
\paragraph{Proof of the local supervision-conflict criterion.}
Consider a GT-prefix parent state $s$ with at least two legal children
and observed next token $y$. Let $a_c$ be the student logit of child $c$,
$p_c=\exp(a_c)/\sum_{j\in\mathcal C(s)}\exp(a_j)$, and let $q$ be the teacher distribution on the same legal-child set. We compare gradients
against the local cross-entropy
$\mathcal L_{\mathrm{SFT}}(s)=-\log p_y$, evaluated using this same softmax.
For
$0<p_y<1$ and $0\le q_y\le1$, differentiating the softmax gives
\begin{align}
\frac{\partial p_y}{\partial a_c}
&=p_y\bigl(\mathbb I[c=y]-p_c\bigr),
\label{eq:cdbd-softmax-derivative}\\
\frac{\partial\mathcal L_{\mathrm{SFT}}}{\partial a_c}
&=p_c-\mathbb I[c=y].
\label{eq:cdbd-sft-gradient}
\end{align}
The mass term in Eq.~\eqref{eq:mass} is a Bernoulli KL divergence given by
\begin{equation}
\lossmass=q_y\log\frac{q_y}{p_y}
 +(1-q_y)\log\frac{1-q_y}{1-p_y}.
\end{equation}
Differentiation with respect to $p_y$ gives
\begin{equation}
\frac{\partial\lossmass}{\partial p_y}
=-\frac{q_y}{p_y}+\frac{1-q_y}{1-p_y}
=\frac{p_y-q_y}{p_y(1-p_y)}.
\end{equation}
Combining this result with Eq.~\eqref{eq:cdbd-softmax-derivative}
yields, for every legal child $c$,
\begin{equation}
\begin{aligned}
\frac{\partial\lossmass}{\partial a_c}
&=\frac{p_y-q_y}{1-p_y}
 \bigl(\mathbb I[c=y]-p_c\bigr)\\
&=-\frac{p_y-q_y}{1-p_y}
 \frac{\partial\mathcal L_{\mathrm{SFT}}}{\partial a_c}.
\end{aligned}
\label{eq:cdbd-mass-gradient-proof}
\end{equation}
In particular,
\begin{equation}
\frac{\partial\mathcal L_{\mathrm{SFT}}}{\partial a_y}=p_y-1,
\qquad
\frac{\partial\lossmass}{\partial a_y}=p_y-q_y.
\label{eq:cdbd-target-gradients}
\end{equation}
It follows that the mass and SFT gradients are \emph{oppositely directed}
in the common-support logit space when $q_y<p_y$, aligned when
$q_y>p_y$, and zero for the mass term when $q_y=p_y$.

For the relation component, when $0<q_y<1$ and at least two non-target legal children are supported, the student distribution conditioned on non-target alternatives is
\begin{equation}
p_c^-=\frac{p_c}{1-p_y}
=\frac{\exp(a_c)}{\sum_{j\ne y}\exp(a_j)},\qquad c\ne y.
\label{eq:cdbd-conditional-softmax}
\end{equation}
This conditional distribution is independent of $a_y$. The standard KL relation factor gives $\lossrel=(1-q_y)\losscond$. Since the teacher target $q^-$ does not depend on student logits,
\begin{equation}
\frac{\partial\losscond}{\partial a_c}=
\begin{cases}0,&c=y,\\p_c^--q_c^-,&c\ne y.
\end{cases}
\label{eq:cdbd-relation-gradient-proof}
\end{equation}
Because $\omega(s)$ in Eq.~\eqref{eq:relation-floor} depends only on detached teacher confidence, the corresponding CDBD relation gradient is $\omega(s)(p_c^--q_c^-)$ for $c\ne y$ and zero for $c=y$. Unlike the standard KL weighting $1-q_y$, this coefficient remains positive when $\omega_{\min}>0$ and $q_y$ approaches one, provided $q^-$ is supported and numerically reliable. Otherwise the conditional relation term is not evaluated. A singleton legal-child set likewise has zero branch loss.

The mass weight $\alpha(s)$ and relation weight $\omega(s)$ are detached. Combining the two components yields the local gradient
\begin{equation}
\nabla_a\mathcal L_{\mathrm{CDBD}}
=\omega(s)\nabla_a\losscond+\alpha(s)\nabla_a\lossmass.
\label{eq:cdbd-gated-gradient}
\end{equation}
For $q_y<p_y$, choosing $\alpha_{\min}<1$ attenuates the opposing mass gradient while preserving the conditional relation gradient. The relation term acts on non-target logits and has zero direct derivative with respect to the target logit. Setting $\alpha_{\min}=1$ and $\omega_{\min}=0$ recovers the local Full-KL term at supported GT-prefix states. The branch loss then applies the $1/|\mathcal C(s)|$ factor in Eq.~\eqref{eq:effective-branch}.

\section{Parameter Sensitivity}
\label{sec:sensitivity}
\begin{figure}[ht]
\centering
\includegraphics[width=0.96\columnwidth]{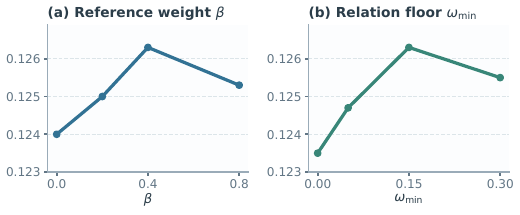}
\caption{Industrial NDCG@10 under one-at-a-time sweeps of (a) the EBS coefficient $\beta$ and (b) the CDBD relation floor $\omega_{\min}$, with the other parameter fixed.}
\Description{For beta equal to 0, 0.2, 0.4, and 0.8, NDCG@10 is 0.1240, 0.1250, 0.1263, and 0.1253. For omega-minimum equal to 0, 0.05, 0.15, and 0.30, NDCG@10 is 0.1235, 0.1247, 0.1263, and 0.1255.}
\label{fig:sensitivity}
\end{figure}

Figure~\ref{fig:sensitivity} examines the sensitivity of SA-OPSD to $\beta$ and $\omega_{\min}$ on Industrial, varying one parameter at a time. As $\beta$ increases from 0 to 0.4, NDCG@10 improves from 0.1240 to 0.1263, before declining to 0.1253 at 0.8. Similarly, increasing $\omega_{\min}$ from 0 to 0.15 improves NDCG@10 from 0.1235 to 0.1263, followed by a decrease to 0.1255 at 0.30.

\end{document}